# A multiagent urban traffic simulation
# Part II: dealing with the extraordinary

Éric Daudé, Pierrick Tranouez, Patrice Langlois

*Abstract*— **In Probabilistic Risk Management, risk is characterized by two quantities: the magnitude (or severity) of the adverse consequences that can potentially result from the given activity or action, and by the likelihood of occurrence of the given adverse consequences.**

**But a risk seldom exists in isolation: chain of consequences must be examined, as the outcome of one risk can increase the likelihood of other risks. Systemic theory must complement classic PRM. Indeed these chains are composed of many different elements, all of which may have a critical importance at many different levels.**

**Furthermore, when urban catastrophes are envisioned, space and time constraints are key determinants of the workings and dynamics of these chains of catastrophes: models must include a correct spatial topology of the studied risk.**

**Finally, literature insists on the importance small events can have on the risk on a greater scale: urban risks management models belong to self-organized criticality theory. We chose multiagent systems to incorporate this property in our model: the behavior of an agent can transform the dynamics of important groups of them.**

*Index Terms*— **Risk management, self-organized criticality, multiagent systems, modeling, simulation.**

## I. INTRODUCTION

SPACE is an important factor of risks situations, not only as a support, but also as an actor in itself of the situation. Risk is space related. In epidemic contexts such as cholera, presence and density of the *vibrio cholera* are dependent both on aquatic reservoir and on the density of population in the environment. Risk has spatial impacts. In environmental context, *flash floods* caused high damages because of their torrential nature and of their high spatial concentrations. Risk management makes tracks in space, and risk sometimes stands to management. In technological context, urban land use and planning reveals some tensions between industrial and residential areas. Risks are multi-layered (world, nations, cities) and imply different kinds of actors, human and non-human. Fight against a possible A *flu* pandemic implies many actors at different levels (World Health Organization, national centers for disease control such as INSERM, local government and doctors) and control measures to reduce risks are both global (air traffic limitation) and local (public services closure). Furthermore, risks are dynamic. In industrial context, one can observe *Domino effect* as an explosion in one site produces secondary accidents in the neighborhood, due to the high concentration of activities.

Risk is defined as a probability of space-time interaction between a source and a target [1]. Four concepts are relevant to this definition and are linked to capture the risk: hazard, intensity, vulnerability and resilience (figure 1).

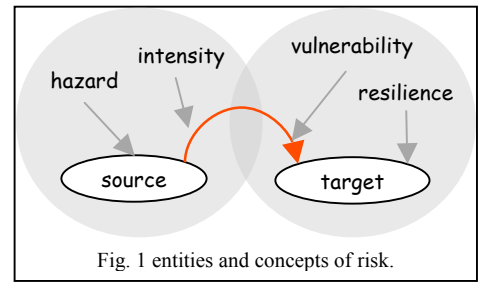

Fig. 1 entities and concepts of risk.

- Hazard represents the occurrence probability of an alteration into the source that could have effects on target: the probability of emergence or re-emergence of a virus for example;
- Intensity is viewed as an output of the source, it depends on the power and duration of the phenomenon and of the involved surface area: the volume and extension of a toxic cloud for example;
- Vulnerability is an input of the target, it measures at the same time the sensitivity of the target to alterations in its environment, and the related

The authors would like to thank the GRR SER and the region Haute-Normandie for the funding of the MOSAIIC program from which this work stems from.
P. Tranouez is with Litis, Rouen University. (e-mail: Pierrick.Tranouez@univ-rouen.fr)
Patrice Langlois and Éric Daudé are with UMR IDEES, Rouen University (email: Patrice.Langlois@univ-rouen.fr and Eric.Daude@univ-rouen.fr)

damages, in term of population or equipments: the probability for a group to panic and to avoid confinement in a technological accident for example;

- Resilience is the capacity for an organization to return gradually an equilibrium state without modifying its final goal: the return time to a normal behavior in a transportation network after a crisis for example.

The main difficulty to characterize risk is the huge amount of interactions that links entities: risk is complex because targets often become sources. And this is all the more true in an urban context characterized by a large number and a great diversity of entities susceptible to be target and source. So space and interactions matter in risk, and they are the two main entrances of our MOSAIIC project.

## II. Dealing with human behaviors in risk context

The MOSAIIC project aims to observe and understand local and global effects of individual behaviors in the dynamic of a transportation network system after an industrial accident. Few researches take into account the behaviors of group or individual when studying the risk at the scale of a city. Physical aspects override the measure of risk and population damage is just a result of these major forces. In this way, intensity of toxic cloud or of earthquake defines buffers that are used to estimate the number of inhabitants and equipment concerned by the event, and then give an estimation of the vulnerability. When human behaviors are considered in risk situation, it is mostly at a very fine scale, for example rooms or building [2], and with the same kind of behavior: panic and escaping [3]. At a global scale, deterministic model are mostly used as they are supposed to be more efficient to describe the mean behavior of individuals, particularly if there is a large number of people concerned. We argue that it is possible, and necessary in some sense, to go beyond this approach.

In risk management, many studies have shown that early stages of the phenomenon are critical on the level of the global damage. It is true with epidemic outbreak when the very few infectious people present in the city can affect, by their individual actions, the course of the epidemic [4]. The same situation can appear when mimetic of panic between some individuals can produce a snow-bowl effect on the entire population. But individual behaviors in risk situations are not limited to panic and escaping behaviors. If one considered Bhopal (1984) or Toulouse (2001) accidents, the number of victims or the resilience of the system have largely increased due to the wondering behavior, curiosity behavior: in some circumstances, people want to see the damage. The aim is then to detect where and in which conditions these bifurcations have a high probability of occurrence in order to prevent them. We then develop a model of simulation in which first, space, as a mediator of interactions, matters. Space is a traffic-oriented network [5]. And second, in which individual behaviors are predominant to explain the dynamic of the vulnerability.

## III. Models of behaviors in extraordinary situations

As described in depth in [5], our model and its resulting simulation builds a transportation graph from GIS data, upon which it create mobile agents modeling vehicles.

These agents enter the network at a controlled random place (their insertion is based on scenarios, they are not necessarily uniformly randomized on the whole network), and try to reach a controlled random destination. To each edge of the network is attributed a *weight*, which combines numerous characteristics of this edge, such as its length, speed limits, number of lanes etc. in a quantification of its attractiveness. This lets our agents the possibility to compute an efficient path from where they are to their destination through Dijkstra's algorithm. The planned trajectory of an agent is then a succession of edges. Once in an edge the agent tries to drive to its end, the next connection, where it will be able to choose the next planned edge.

Our mobile agents then drive to their destination, interacting one with another, as their speed, length, driving brashness etc. are considered at each step. Furthermore, these agents can adapt their goals to what they perceive of the traffic, using different methods to choose other paths to reach their destination.

MOSAIIC is concerned by situations where contextual mobility can occur and can diffuse or have large consequences on the global circulation. We call *contextual mobility* a mobility which is associated to short-range goals (to avoid a crowd) and whose result differs from the initial planning (to change a destination). We will now consider an urban industrial accident. This accident has a finite extension area and well-determined intensity, represented by a buffer. Inside this buffer, a proportion of people, related to intensity, want to escape. Outside this buffer, behaviors are less reactive. Some want to escape, others want to see and for others "show must go on", and they want to follow their way. We have then defined different kinds of behaviors and methods related to these different goals:

- *Chicken behavior*: the goal is to find the opposite direction of the source (the buffer), and to drive following this way;

- *Bystander behavior*: the goal is to find the source of danger and to go there. If agent is already in the place, then he stays here;

- *Pragmatic behavior*: here the agent selects a new destination in the network and tries to reach it. This behavior simulates the fact that some people will want to reach their children at school or husband or wife at their working place;

- *Wandering behavior*: there is no goal, this behavior is the sign of distress. At each time step, just select a road and go there.

- *Roadrunner behavior*: this method consists in always selecting the less congested road and to go there. This method can be connected to the Chicken or Bystander behavior;

- *Sheep behavior*: here agent follows the crowd whatever the direction.

We will now present implementations of these behaviors.

## IV. SIMULATION OF BEHAVIORS

We will discuss here how the behaviors themselves can be implemented, not why or when one or the other will be chosen.

### A. Behaviors classification

In order to implement them, we will distinguish three *categories* of behavior: global, planar and local. These categories are based on the actual behavior, and not on the motivations behind it.

A *global* behavior is one that makes a reasoning about the road network. *Pragmatic* behavior will probably fall in this category: the agent will try to find a good path to his newly decided destination using his knowledge of the network. *Bystander* can also fall here.

A *planar* decision also chooses a destination but tries to reach it using orientation as if no roads existed, as if the vehicle was on an open plan. Of course the network will offer constraints, but a general cardinal like direction will guide the agent. *Chicken* and possibly B*ystander* will fall in this category. This means there are two sub-behaviors in by standing.

A *local* decision is one based on local-only data: *Wandering*, *Roadrunner* and *Sheep* will fall there.

### B. Class implementations

*Global* behaviors are implemented in the agents to allow them to reach their initial destination.

*Local* require little complexity. *Wandering* is trivial, *Roadrunner* and *Sheep* differ only by the sign of their optimization. We also implemented a simple anti-loop measure: *Roadrunners* for example will choose the less congested road unless they already went recently through this one.

*Planar* require the ability to choose an edge out of a node based on a global direction. Depending on what the modeler desires, he can choose a distance from the current road intersection, and the agent will choose the intersection at less than the selected distance (expressed in Euclidean distance or number of edges in a path leading to it) that is the closest to the desired direction. An anti-loop measure can be added.

### C. Examples of simulation

The behaviors previously described are ways of coping with an extraordinary situation. Most urban important accidents will have their consequences felt locally at the beginning, before it spreads. The evolution of the perturbation will be like waves spreading from the initial locus outward. If the extraordinary behaviors are the waves, the metaphorical medium of this propagation is the ordinary traffic flow. We therefore need a sophisticated modeling of the day-to-day activities of vehicles in an urban agglomeration. We described this model in [5].

With the simulation of ordinary traffic, one can see in figure n°2 an example of the distribution of vehicles in the main roads of the city.

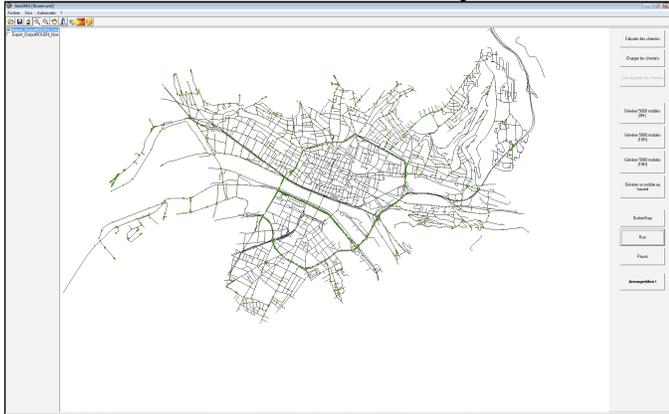

Fig. 2: An example of traffic in a town before an accident occurs.

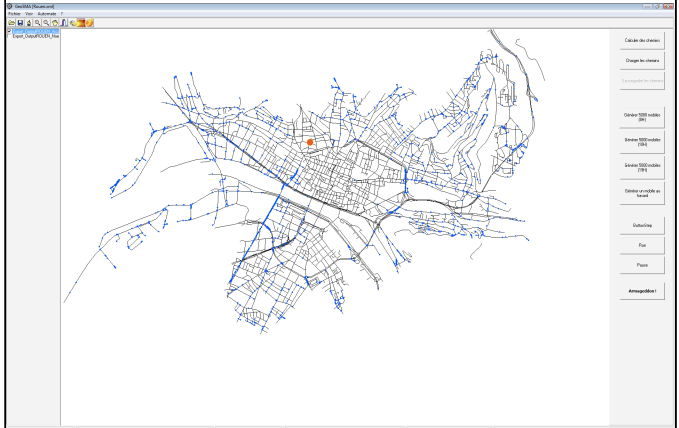

Fig. 3: The same traffic after the accident occurs (red circle). The agents are all adopting *Chicken behavior* (in blue).

Starting at this point, we generate an event in the city that is supposed to represent an accident. This event, for the purpose of the simulation, is perceived by all individuals and is considered as a repulsive event. In figure n°3, this event is a mouse-click event located by the user without any consideration about the reality of the area. As agents perceived the impact zone (in fact XY coordinates), they all change their planned trajectory. Once in a crossroad, all mobiles pick out the *Chicken behaviors* and compute their new XY position using:

$$XY_{(t+1)} = \text{Best value (min (VxVy explosion - VxVy edge))}$$

This escaping behavior is for instance not applied in concurrence with any other mobility strategies or tactical behavior: they have not the possibility to avoid traffic jam or loops. The main effect of the general application of this rule is purely the draining of the transportation network. Of course this "Hollywood" scenario is not relevant in reality but let us test implemented mechanism.

Vulnerability increases when a certain quantity of actors changed their dynamics of mobility, mainly after a shift in their goals. Beliefs, desires and goals are then important to take into account in this kind of model.

## V. DISCUSSION

We have defined methods modeling mobility itself, but we now need to model the decision processes for picking or switching between these methods. In an ordinary situation, people follow their own planning and most of the time never deviate of their schedule. But how to justify and explain the fact that in some circumstances, people shift from one behavior to another, from an ordinary behavior to one of the extraordinary described here such as *Sheep* or *Roadrunner* ?

Each agent can be seen as a cognitive agent, where motivation is important in the act of mobility. Motivation is "life dependant", and "contextual dependant": we can say that there is a path dependence of the individual motivations, where the present and future is mainly conditioned by the past; and that sometimes motivations, in a short spatiotemporal range, depart and express something really different. This last conception can be seen as the result of processes such as adaptation, evolution, archaic instinct and so on and so forth. In our debate, this is linked to the fact that people are able to change their plans and that they do not want to keep going to the previously planned destination.

We are not fathoming here the psychological processes that lead from one objective to another, the main point is the result of such behaviors. We have to think of a way to sum-up individual intelligence by simple processes.

Architecture such as Beliefs - Desires - Intentions (BDI) [7] is probably well adapted to this kind of cognitive agent.

- Belief here represents the schedule in normal

situation, information about environment (other mobile agents and road network) and attributes of agents that can describe risk culture, sociability, tolerance level etc. Belief is subject to uncertainty and error. In our model, Belief play a role as representation of industrial accident and its dynamic is important in human behaviors. Both the spatial and temporal distance of the accident can modulate how it is perceived.

- Desires are goals assigned to the agent, they are influenced by beliefs. Desires represent in our model different points to reach in space.

- Intentions represent the priority for an agent to achieve goals: it can represent a sort of utility function where each element is a point in space.

f(g)={X1Y1; X2Y2;...XiYj}

Intentions are then both a goal and a list of goals.

In a disaster situation, agent receives different kind of information (Beliefs). If following his curiosity or instincts is a predominant goal of the agent (Desires), or fear or cupidity etc. then he will permute elements of his utility function, and even add new elements in order to plan new actions (Intentions).

The main question is then: is it necessary to have a good knowledge of people desires to simulate crowd dynamics and vulnerability of transportation network? In other words, what level of detail is needed in the modeling of individual agent to accurately model the beahvior of a crowd of them ?